\documentclass{article}

\usepackage{amsmath,amsthm,amsfonts,amssymb,natbib,graphicx,mathtools, enumerate}
\usepackage{float}
\usepackage[margin=1.3in]{geometry}
\usepackage[blocks, affil-it]{authblk}
\usepackage{bbm}
\newtheorem*{theorem*}{Theorem}

\newtheorem{result}{Result}

\newtheorem{corollary}{Corollary}

\newcommand{\ha}{\hat a}
\newcommand{\brac}[1]{\left(#1\right) }
\newcommand{\E}{\mathbb{E}}
\newcommand{\Prob}{\mathbb{P}}
\newcommand{\indc}[1]{\mathbbm{1}_{\{#1\}}}
\newcommand{\obj}{\textnormal{obj}}
\newcommand{\yn}{y_{\textnormal{new}}}
\newcommand{\xn}{x_{\textnormal{new}}}
\newcommand{\R}{\mathbb{R}}
\newcommand{\norm}[1]{\|#1\|}
\newcommand{\prox}{\textnormal{prox}}
\newcommand{\avg}[1]{\langle #1\rangle}

\title{Exact high-dimensional asymptotics for Support Vector Machine}
\author{Haoyang Liu}
\affil{
	University of Chicago
}

\begin{document}

\maketitle

\begin{abstract}
The Support Vector Machine (SVM) is one of the most widely used classification methods. In this paper, we consider the soft-margin SVM used on data points with independent features, where the sample size $n$ and the feature dimension $p$ grows to $\infty$ in a fixed ratio $p/n\rightarrow \delta$. We propose a set of equations that exactly characterizes the asymptotic behavior of support vector machine. In particular, we give exact formulas for (1) the variability of the optimal coefficients, (2) the proportion of data points lying on the margin boundary (i.e. number of support vectors), (3) the final objective function value, and (4) the expected misclassification error on new data points, which in particular implies the exact formula for the optimal tuning parameter given a data generating mechanism. We first establish these formulas in the case where the label $y\in\{+1,-1\}$ is independent of the feature $x$. Then the results are generalized to the case where the label $y\in\{+1,-1\}$ is allowed to have a general dependence on the feature $x$ through a linear combination $a_0^Tx$. These formulas for the non-smooth hinge loss are analogous to the recent results in \citep{sur2018modern} for smooth logistic loss. Our approach is based on heuristic leave-one-out calculations. 
\end{abstract}

\section{Introduction}\label{sec:intro}
The Support Vector Machine (SVM) is one of the most standard methods for data classification (\citep{vapnik2013nature,vapnik1998support}). The standard theoretical analysis of SVM is formulated in the framework of statistical learning theory (see for example \citet{vapnik2000bounds}). This type of analysis has the advantage of being general and flexible, in the sense that it poses rather weak conditions on the data-generating mechanism. On the other hand it usually depends on different upper bounds on complexity measures and is not exact. In this paper, we thus study a more restrictive classification setting where the different features of the data points are assumed to be independent. This allows us to provide an analysis for the SVM that is asymptotically exact when the dimension $p$ of the feature space and the sample size $n$ grow together in a fixed ratio.

There is a large body of theoretical works in the setting of high dimensional regression and classification under the asymptotic setting where $p$ and $n$ grow in proportion (\citep{sur2018modern,bayati2011dynamics,bean2013optimal,donoho2009message,donoho2011noise,donoho2016high,el2013robust,huang2017asymptotic,javanmard2013state,karoui2013asymptotic,sur2017likelihood}). In particular, \citep{huang2017asymptotic} also studies the asymptotic properties of the SVM, where they model the feature conditional on the label and consider a spiked model for the features. In comparison, we model the label conditional on the feature and the label is allowed to have a general dependence on the feature through a linear combination.

\subsection{Problem formulation}
Consider the problem of classifying $n$ data points $(x_i,y_i)$ for $i=1,\dots,n$, where $x_i\in \R^p$ and $y_i\in\{+1,-1\}$. The data points $(x_i,y_i)$'s are assumed to be generated i.i.d. for all $i$'s with $x_i\overset{\textnormal{i.i.d}}{\sim} N(0,I_p)$\footnote{The result is expected to hold as long as $x_i(j)$ are generated i.i.d with some moment condition.}. With the notation $m_+=\max(0,m)$, the soft-margin support vector machine solves the following minimization problem:
\[\ha=\arg\min_a \sum_{i=1}^n \brac{1-\frac{y_ix_i^Ta}{\sqrt{p}}}_++\lambda \sum_{j=1}^p a^2(j),\]
with some penalty parameter $\lambda>0$. Throughout the paper, we work under the asymptotics that $p\rightarrow \infty, n\rightarrow \infty, p/n\rightarrow \delta$. Given a fixed $\delta>0$ and a fixed $\lambda>0$, we ask the following questions in the $p/n\rightarrow \delta$ limit:
\begin{itemize}
	\item[1] What is the distribution of the coefficients $\ha(1),\dots,\ha(p)$?
	\item[2] What is the distribution of the linear combinations $\frac{y_1x_1^T\ha}{\sqrt{p}},\dots,\frac{y_nx_n^T\ha}{\sqrt{p}}$? In particular, what is the proportion of the data points lying on the margin boundary?
\end{itemize}
And as applications of the knowledge above,
\begin{itemize}
	\item[3] What is the final objective value? In other words, what is the typical value of
	\[\frac{1}{n}\brac{\sum_{i=1}^n \brac{1-\frac{y_ix_i^T\ha}{\sqrt{p}}}_++\lambda \sum_{j=1}^p \ha^2(j)}\textnormal{?}\]
	\item[4] What is the expected misclassification rate on a new data point? In other words, what is the typical value of
	\[\Prob(\yn\xn^T\ha<0),\]
	and what is the optimal tuning parameter $\lambda$ given a data-generating mechanism?
\end{itemize}
In the following sections, we provide exact answers to these questions. The global null case are considered first, where the label $y\in\{+1,-1\}$ is independent of the feature $x$. Then the results are generalized to the signaled case, where the label $y\in\{+1,-1\}$ is allowed to have a general dependence on the feature $x$ through a linear combination $a_0^Tx$. Our approach is similar to the one adopted in \citet{el2013robust}, where heuristic leave-one-out calculation is used. The correctness of the analytic formulas and its accuracy in finite sample is verified through various simulations in Section~\ref{sec:simulations}. We don't pursue a rigorous treatment in this paper. 

\subsection{Notations}
Throughout the paper, we use $Z$ to denote a random variable following standard normal distribution $N(0,1)$. We use $i$ to denote an index in $[n]$, and $j$ to denote an index in $[p]$. The limit $\lim_{n\rightarrow \infty, p\rightarrow \infty, p/n\rightarrow \delta}$ will be abbreviated as $\lim_{n\rightarrow \infty}$ or $\lim_{p\rightarrow \infty}$ depending on the context.

\section{The SVM under the global null}\label{sec:globalnull}
In this section we assume that the label $y$ is independent of the feature $x$ and follows a uniform distribution on $\{+1,-1\}$. Given a fixed $\delta>0$ and a fixed $\lambda> 0$, define the following set of equations on $\gamma$ and $\sigma^2$ under the constraints $\gamma>0,\sigma>0$:
\begin{align}
\label{eq:main}
&(2\lambda\gamma-1)\delta+1=\Prob(\sigma Z\leq 1-\gamma)+\Prob(\sigma Z\geq 1)\\
\nonumber
&\frac{\sigma^2\delta}{\gamma^2}=\Prob(\sigma Z\leq 1-\gamma)+\E\left(\frac{1-\sigma Z}{\gamma}\right)^2\indc{1-\gamma\leq \sigma Z \leq 1}.
\end{align}
Here the probability and expectation are over the randomness of a standard normal random variable $Z\sim N(0,1)$. It is expected that equation~(\ref{eq:main}) has a unique finite solution under the constraint $\gamma>0,\sigma>0$ for all $\lambda>0,\delta>0$. Granted this, we use $(\gamma_*, \sigma_*)$ to denote the solution to it. Further define the following function given $\gamma$:
\[f_\gamma(m)=\min(\max(1,m),m+\gamma).\]
Then the following results show that $(\gamma_*,\sigma_*)$ exactly characterizes the behavior of the SVM under the global null asymptotically.
\begin{result}\label{thm:1}
	Under regularity conditions\footnote{\label{note:1}It is expected that a Lipschitz-type condition similar to the one used in \citet{bayati2011dynamics} is enough. We don't pursue a rigorous treatment here.} on a function $\psi$, we have almost surely
	\[\lim_{p\rightarrow \infty} \frac{1}{p}\sum_{j=1}^p\psi(\ha(j))=\E \psi(\sigma_* Z).\] 
\end{result}
\begin{result}\label{thm:2}
	Under regularity conditions\textsuperscript{$\ref{note:1}$} on a function $\psi$, we have almost surely
	\[\lim_{n\rightarrow \infty} \frac{1}{n}\sum_{i=1}^n\psi\brac{\frac{y_ix_i^T\ha}{\sqrt{p}}}=\E \psi\brac{f_{\gamma_*}(\sigma_* Z)}.\]
\end{result}
\begin{result}\label{thm:3}
	Let $n_b$ be the number of data points on the margin boundary, i.e.
	\[n_b=\textnormal{\#}\left\{i:\frac{y_ix_i^T\ha}{\sqrt{p}}=1\right\}.\]
	Then almost surely
	\[\lim_{n\rightarrow \infty} \frac{n_b}{n}=(1-2\lambda\gamma_*)\delta.\]
\end{result}
\noindent A heuristic derivation of equation~(\ref{eq:main}) is given in the appendix. With these results at hand, we can interpret the meaning of $(\gamma_*,\sigma_*)$. Due to Result~\ref{thm:1}, the coefficients $\ha(j)$'s behave as independent samples from $N(0,\sigma_*^2)$. Due to Result~\ref{thm:3}, $\gamma_*$ is connected to the proportion of points lying on the margin boundary.  Define $\obj_n$ to be the normalized objective value, i.e.
\[\obj_n=\frac{1}{n}\brac{\sum_{i=1}^n \brac{1-\frac{y_ix_i^T\ha}{\sqrt{p}}}_++\lambda \sum_{j=1}^p \ha^2(j)}.\]
Then Result~\ref{thm:1} and Result~\ref{thm:2}  together imply that
\begin{corollary}\label{cor:obj}
	Almost surely
	\[\lim_{n\rightarrow \infty}\obj_n= \E \brac{1-f_{\gamma_*}(\sigma_* Z)}_++\lambda\delta\sigma_*^2.\]
\end{corollary}
\noindent To get some intuition about the solution to the system of equation~(\ref{eq:main}), we discuss two special cases, the small penalty case and the large penalty case.
\paragraph{The small penalty case} In this case, we fix a $\delta>0$ and consider the behavior of the solution $(\gamma_*,\sigma_*)$ as $\lambda$ approaches $0$ from above. Then there is a phase transition point at $\delta=0.5$. When $\delta<0.5$, the solution $(\gamma_*,\sigma_*)$ stays finite when $\lambda$ approaches $0$. However, when $\delta>0.5$, the solution $\gamma_*$ goes to infinity as $\lambda$ approaches $0$. This is best understood if we consider equation~(\ref{eq:main}) with $\lambda=0$.
\begin{align}\label{eq:gamma0}
&1-\delta=\Prob(\sigma Z\leq 1-\gamma)+\Prob(\sigma Z\geq 1)\\
\nonumber
&\frac{\sigma^2\delta}{\gamma^2}=\Prob(\sigma Z\leq 1-\gamma)+\E\left(\frac{1-\sigma Z}{\gamma}\right)^2\indc{1-\gamma\leq \sigma Z \leq 1}.
\end{align}
\noindent An inspection of equation~(\ref{eq:gamma0}) shows that there is no solution when $\delta>0.5$. This is easy to understand: when $\delta>0.5$, it is a classical result (\citet{cover1965geometrical}) that the data points from two classes can be separated perfectly by a hyperplane with high probability. Therefore, when the penalty gets small, the problem become ill-posed and the objective function would be trivially close to $0$.
\paragraph{The large penalty case} When $\lambda$ is large, both $\gamma_*$ and $\sigma_*$ would be small. And by keeping the leading order term, equation~(\ref{eq:main}) becomes:
\begin{align*}
&2\lambda\gamma \approx 1\\
&\sigma^2\delta\approx \gamma^2.
\end{align*}
And therefore we have $\gamma_*\approx \frac{1}{2\lambda}$ and $\sigma\approx \frac{1}{2\lambda\sqrt{\delta}}$. By Result~\ref{thm:3}, this implies that when $\lambda$ is large, the proportion of data points lying on the margin boundary would be close to $0$ as expected. Indeed from Corollary~\ref{cor:obj} and equation~(\ref{eq:main}) we have:
\[\lim_{n\rightarrow \infty} \frac{n_b}{n}=(1-2\lambda\gamma_*)\delta=\Prob(1-\gamma_*\leq \sigma_* Z\leq 1).\]
This implies that the proportion of data on the margin boundary would decrease exponentially with $\lambda$.

\section{The SVM with signaled data}\label{sec:linearsignal}
In this section we assume that there is a ground truth direction $a_0$ with $\norm{a_0}^2=p$ such that $y$ is generated depending on $a_0^Tx/\sqrt{p}$. In particular, we assume
\begin{align*}
&x\sim N(0,I_p),\\
&\Prob (y=1|x)=\ell \brac{\frac{a_0^Tx}{\sqrt{p}}},
\end{align*}
for some function $\ell$ taking value in $[0,1]$. Throughout this section, we use $V$ to denote a random variable that has the same distribution as $ya_0^Tx/\sqrt{p}$. Given a fixed $\ell$ and thus a fixed distribution of $V$, we let $p,n$ grows to infinity with $p/n\rightarrow \delta$.

Now we describe the system of equations that characterizes the behavior of SVM in the signaled case. Given an $\alpha$, let $(\gamma_\alpha,\sigma_\alpha^2)$ be the solution to the following system of equations on $(\gamma,\sigma)$ under the constraints $\gamma>0,\sigma>0$ whenever they exist.
\begin{align}
\label{eq:mainsignal}
&(2\lambda\gamma-1)\delta+1=\Prob(\alpha V+\sigma Z\leq 1-\gamma)+\Prob(\alpha V+\sigma Z\geq 1)\\\nonumber
&\frac{\sigma^2\delta}{\gamma^2}=\Prob(\alpha V+\sigma Z\leq 1-\gamma)+\E\left(\frac{1-(\alpha V+\sigma Z)}{\gamma}\right)^2\indc{1-\gamma\leq \alpha V+\sigma Z \leq 1}.
\end{align}
The randomness is over an independent pair of $V$ and $Z\sim N(0,1)$. Then the optimal $\alpha$ is determined through the following formula:
\begin{equation}\label{eq:alpha}
\alpha_*=\arg\min_{\alpha} \E \brac{1-f_{\gamma_\alpha}(\alpha V+\sigma_\alpha Z)}_++\lambda\delta(\sigma_\alpha^2+\alpha^2).
\end{equation}
Again the randomness is over an independent pair of $V$ and $Z\sim N(0,1)$. Given the definition of $\alpha_*$, we further define 
\[(\gamma_*,\sigma_*)=(\gamma_{\alpha_*},\sigma_{\alpha_*}).\]
We expect that with $\lambda>0$, under regularity conditions on the distribution of $V$ (equivalently on the function $\ell$), the triple $(\alpha_*,\gamma_*,\sigma_*)$ by the above definition exists uniquely. Granted this, we have the following results in the signaled case analogous to the results in Section~\ref{sec:linearsignal}.
\begin{result}\label{thm:4}
	Under regularity conditions\footnote{\label{note:2}Again, it is expected that a Lipschitz-type condition for $\psi$ similar to the one used in \citet{bayati2011dynamics} is enough. We don't pursue a rigorous treatment here.} on a function $\psi$ and the model $\ell$, we have almost surely
	\[\lim_{p\rightarrow \infty} \frac{1}{p}\sum_{j=1}^p\psi(\ha(j)-\alpha_*a_0(j))=\E \psi(\sigma_* Z).\] 
\end{result}
\begin{result}\label{thm:5}
	Under regularity conditions\textsuperscript{$\ref{note:2}$}  on a function $\psi$ and the model $\ell$, we have almost surely
	\[\lim_{n\rightarrow \infty} \frac{1}{n}\sum_{i=1}^n\psi\brac{\frac{y_ix_i^T\ha}{\sqrt{p}}}=\E \psi\brac{f_{\gamma_*}(\alpha_*V+\sigma_* Z)}.\]
\end{result}
\begin{result}\label{thm:6}
	Let $n_b$ be the number of data points on the margin boundary, i.e.
	\[n_b=\textnormal{\#}\left\{i:\frac{y_ix_i^T\ha}{\sqrt{p}}=1\right\}.\]
	Then almost surely
	\[\lim_{n\rightarrow \infty} \frac{n_b}{n}=(1-2\lambda\gamma_*)\delta.\]
\end{result}
\begin{result}\label{thm:7}
	For the misclassification error at a new data points, we have almost surely
	\[\lim_{n\rightarrow \infty}\Prob_{(\xn,\yn)}(\yn\xn^T\ha<0)=\Prob_{(V,Z)}(\alpha_*V+\sigma_*Z<0).\]
	where the probability on the left hand side is over the randomness of a new sample generated from the same model given by $\ell$ and $a_0$, and the probability on the right hand side is over the randomness of the independent pair $(Z,V)$. Note that $\Prob(\yn\xn^T\ha<0)$ is itself a random variable depending on the randomness of $\ha$.
\end{result}
\noindent With these results at hand, we can interpret the solution $(\alpha_*,\gamma_*,\sigma_*)$. The parameter $\alpha_*$ is such that $\ha$ fluctuates around $\alpha_*a_0$. The parameter $\sigma_*$ can be viewed as the standard deviation of $a(j)$'s after subtracting their mean $\alpha_*a_0(j)$'s. $\gamma_*$ is still connected to the proportion of data points on the margin boundary. By comparing the form of equation~(\ref{eq:mainsignal}) and Corollary~\ref{cor:obj}, we see that equation~(\ref{eq:mainsignal}) just states that the $\alpha_*$ should be such that the objective value is at its minimum. Due to Result~\ref{thm:7}, the misclassification rate is simply the probability that the random variable $\alpha_*V+\sigma_*Z$ is negative. For $a_0$ to be a genuine signal, $\ell\brac{a_0^Tx/\sqrt{p}}$ is large if $a_0^Tx/\sqrt{p}$ is large and vice versa. Therefore, we expect $V$ to follow a distribution that is positive most of the time. Therefore the misclassification error will be smaller than $1/2$, which is the misclassification error on a new data point under the global null. 

\paragraph{The optimal tuning parameter} Result~\ref{thm:7} allows us to determine the optimal penalty parameter $\lambda$. Given a model $\ell$ and the induced random variable $V$, the optimal $(\alpha_*,\gamma_*,\sigma_*)$ as defined before are functions of $\lambda$, which we now denote as $(\alpha_*(\lambda),\gamma_*(\lambda),\sigma_*(\lambda))$, then the optimal tuning parameter is just
\begin{equation}\label{def:optlambda}
\lambda_*=\arg\min_{\lambda>0}\Prob\left(\alpha_*(\lambda)V+\sigma_*(\lambda)Z<0\right). 
\end{equation}
If there is a $\lambda$ such that $\alpha_*(\lambda)$ is much larger than $\sigma_*(\lambda)$, then the misclassification error would be close to $\Prob(V<0)$,
which is the error of the oracle classifier $\hat y=\textnormal{sign}(a_0^Tx)$. Of course under a high dimensional asymptotics this is typically not the case and $\alpha_*(\lambda)$ and $\sigma_*(\lambda)$ are typically large or small in the same time. And the $\lambda^*$, as defined in equation~\eqref{def:optlambda} , achieves the optimal trade-off between increasing $\alpha_*(\lambda)$ and decreasing $\sigma_*(\lambda)$. Now we consider some special cases of $\ell$.
\begin{itemize}
	\item If $\ell\equiv 1/2$. Then our model is just:
	\[\Prob (y=1|x)=1/2.\]
	Then we have $V\sim N(0,1)$. In this case, an inspection of the equations shows $\alpha_*=0$. Plugging this in, we get back the set of equations in the  global null case. 
	\item If $\ell$ is the logistic function, we have
	\[\Prob(y=1|x)=\frac{1}{1+\exp\left(-c\cdot x^Ta_0/\sqrt{p}\right)}.\]
	In this case $V$ has density function proportional to $\frac{\exp(-x^2/2)}{1+\exp(-cx)}$. We will solve our equations for specific cases in Section~\ref{sec:simulations}.
	\item If $\ell$ is an indicator function of whether its argument is positive, we have
	\[\Prob(y=1|x)=\indc{x^Ta_0\geq 0}.\]
	In this case, $V$ has the same distribution as $|Z|$ where $Z\sim N(0,1)$. We will solve our equations for specific cases in Section~\ref{sec:simulations}.
\end{itemize}

\section{Empirical results}\label{sec:simulations}
In this section we conduct simulations to verify the finite-sample accuracy of the analytic formulas.
\subsection{The global null case}
\begin{figure}
	\centering
	\includegraphics[width=5cm,height=5cm]{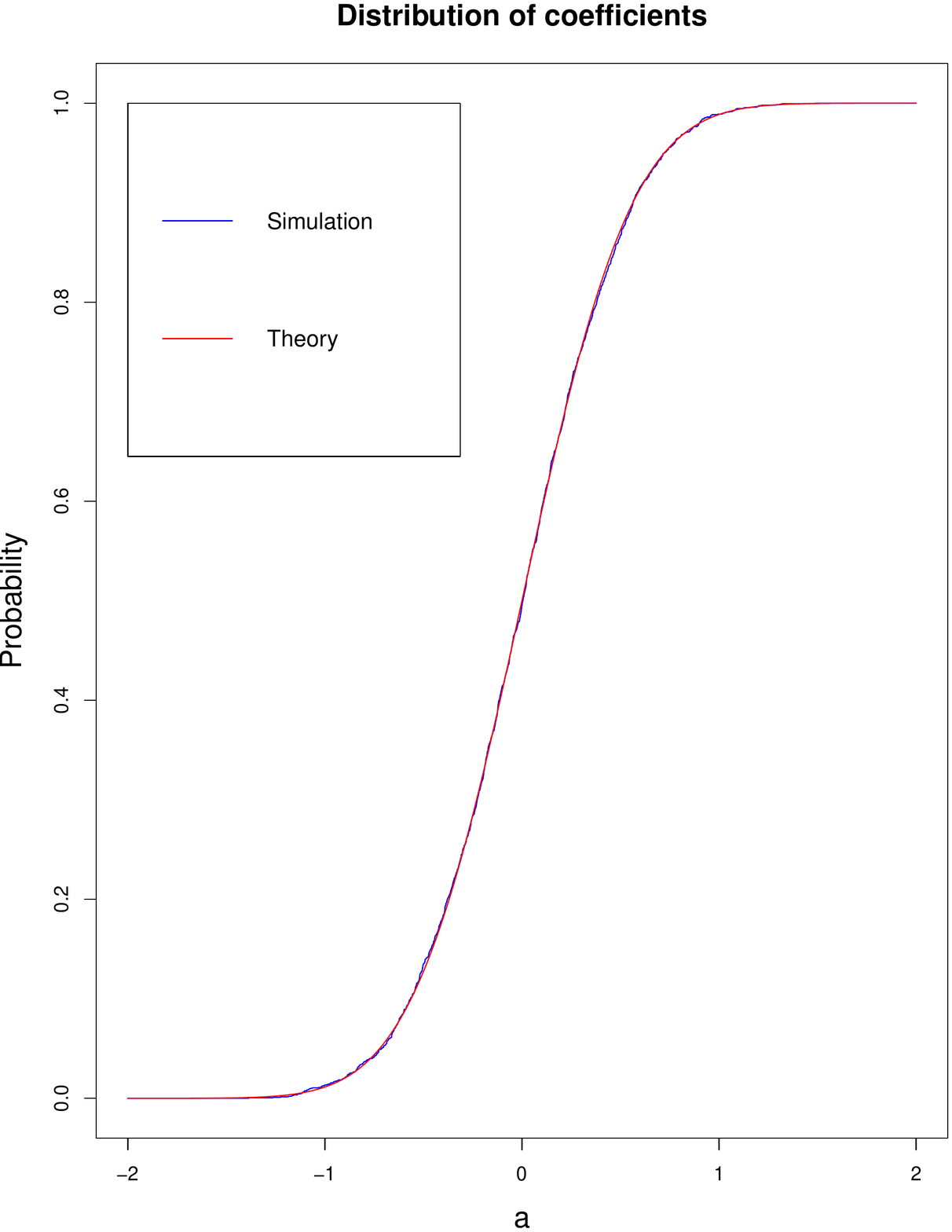}
	\includegraphics[width=5cm,height=5cm]{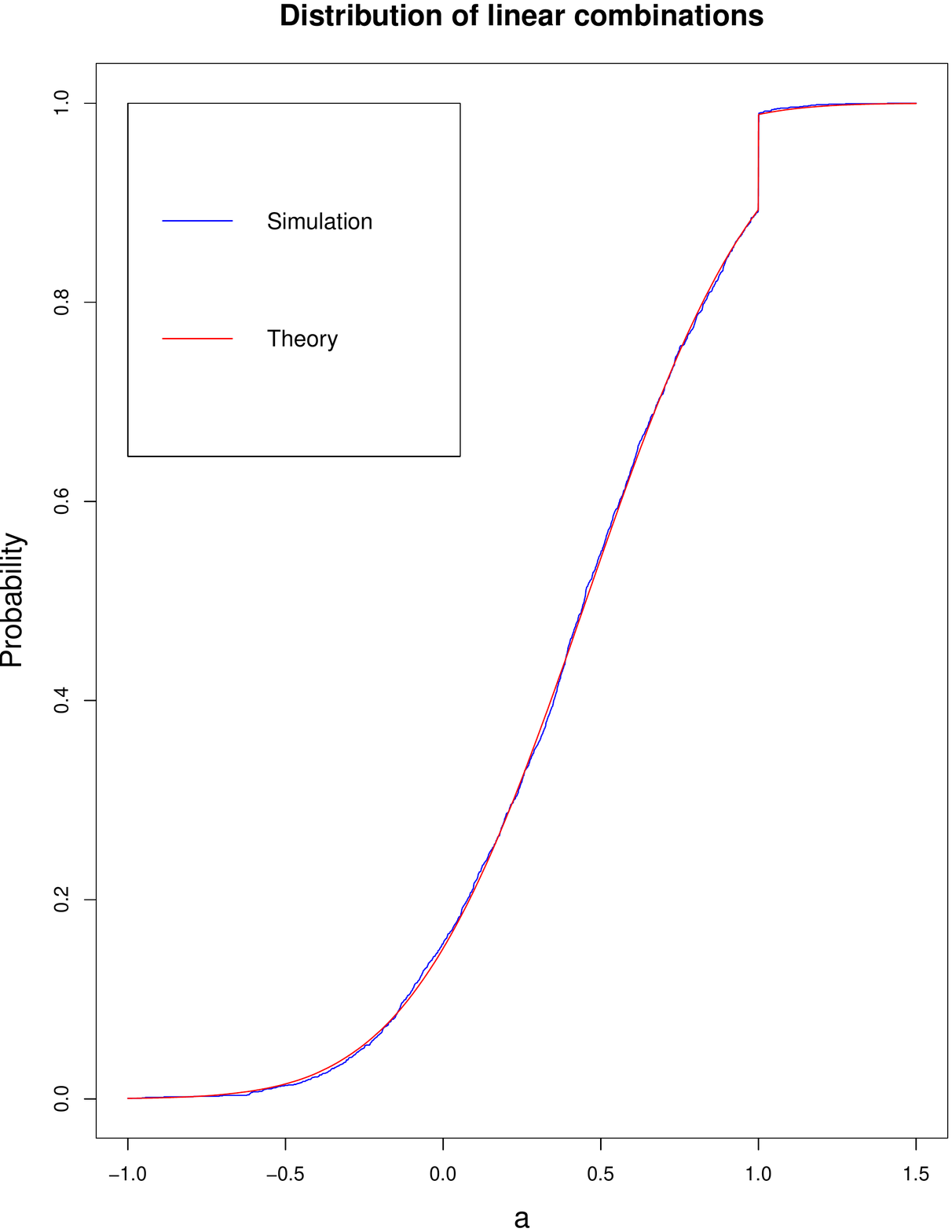}
	\caption{Comparison of simulation result and the analytic formula for the distribution. The first plot is for the coefficients $\ha(j)$'s and the second plot is for the linear combinations $L_i$'s. In both plots, the blue line is the empirical distribution obtained by simulation and the red lines is the theoretic prediction.}
	\label{fig:nosignal12}
\end{figure}

In the global null case we generate our data with $p=2000,n=2000$ and we run SVM with $\lambda=1$. This corresponds to $\delta=\frac{p}{n}=1$. Solving the system of equations~(\ref{eq:main}) gives approximately
\[\gamma_*=0.45,\sigma_*=0.44.\]
Then our analytic formulas predicts the following:
\begin{itemize}
	\item $\ha(j)$'s approximately follow a $N(0,\sigma_*^2)$ distribution.
	\item The linear combinations $L_i=\frac{x_i^T\ha}{\sqrt{p}}$'s approximately follow the same distribution as $f_{\gamma_*}(\sigma_* Z)$.
\end{itemize}
These assertions are examined in figure~\ref{fig:nosignal12}, where the empirical cumulative distribution function are drawn together with the theoretical cumulative distribution predicted by our theory. In the second plot, the fact that the two jumps at $1$ match each other means that our formula for the proportion of data points lying on the margin boundary is also accurate. These plots show that the analytic formulas are very accurate even when the sample size and the dimension are of several thousand. Now we turn to the signaled case.

\subsection{The signaled case}
\subsubsection{The logistic model}
\begin{figure}
	\centering
	\includegraphics[width=5cm,height=5cm]{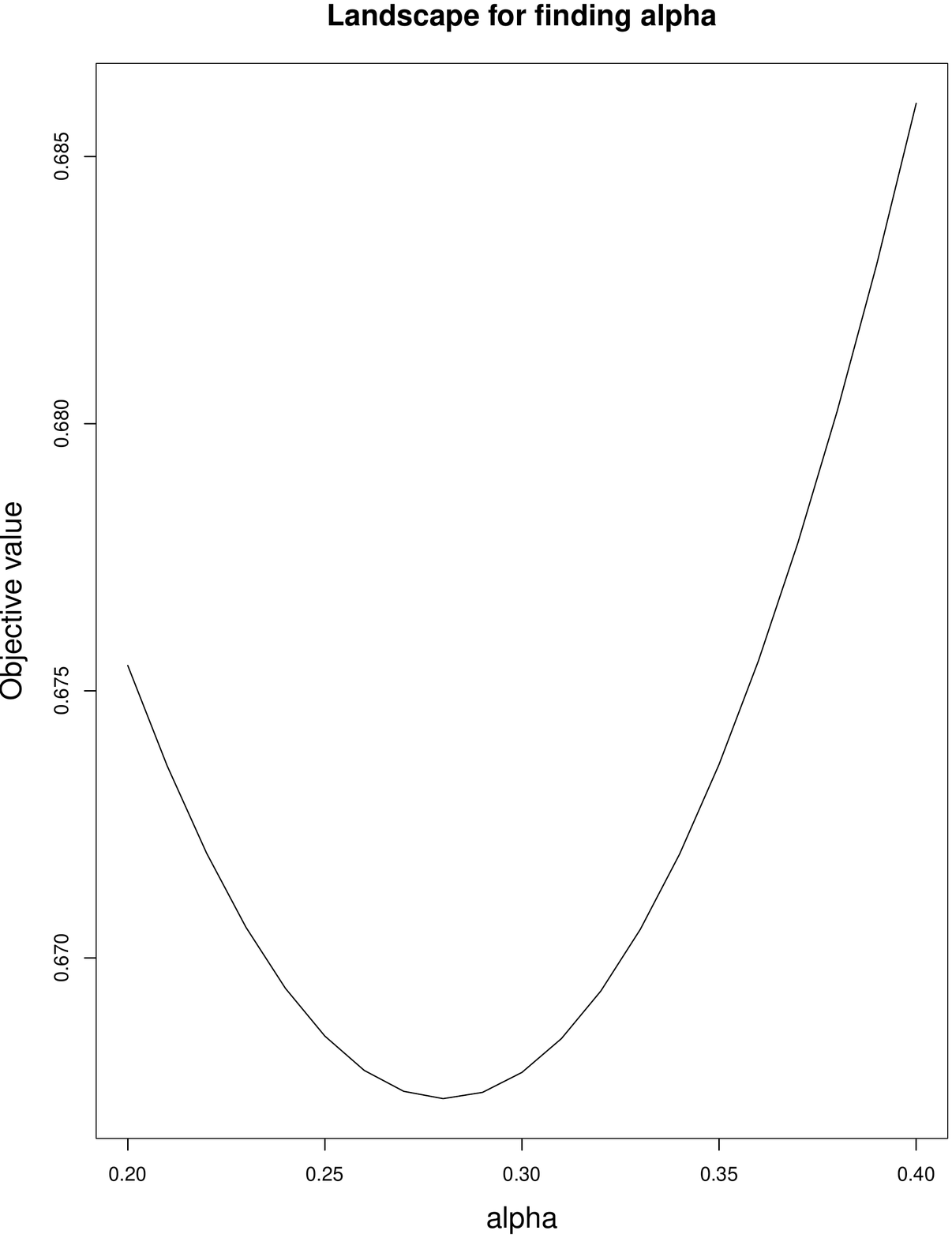}
	\includegraphics[width=5cm,height=5cm]{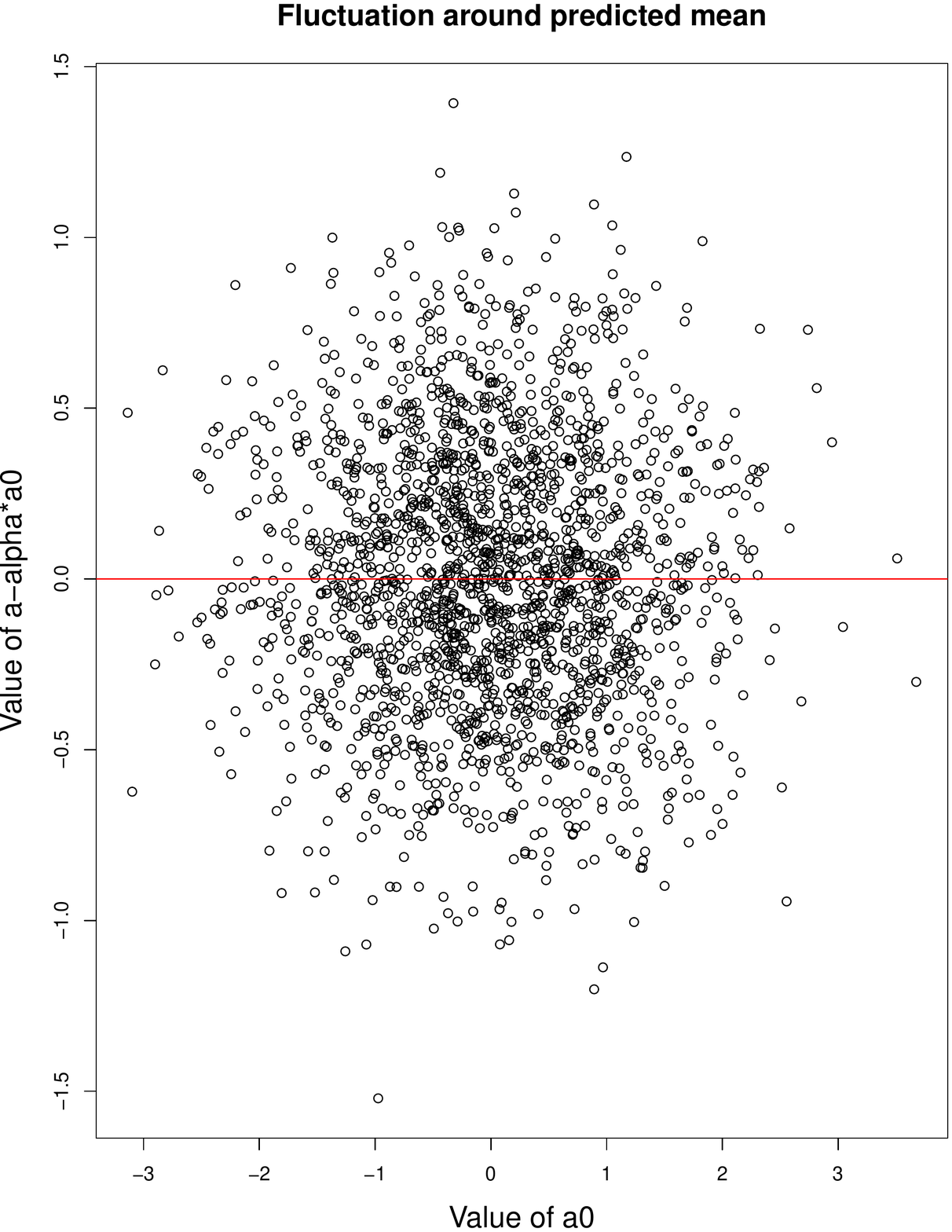}
	\caption{The first plot shows the landscape for the minimization problem in equation~(\ref{eq:alpha}). The second plot shows the values of $\ha(j)-\alpha_*a_0(j)$'s against the values of $a_0(j)$'s. The red line in the second plot is $y=0$.} 
	\label{fig:together}
\end{figure}
\begin{figure}
	\centering
	\includegraphics[width=5cm,height=5cm]{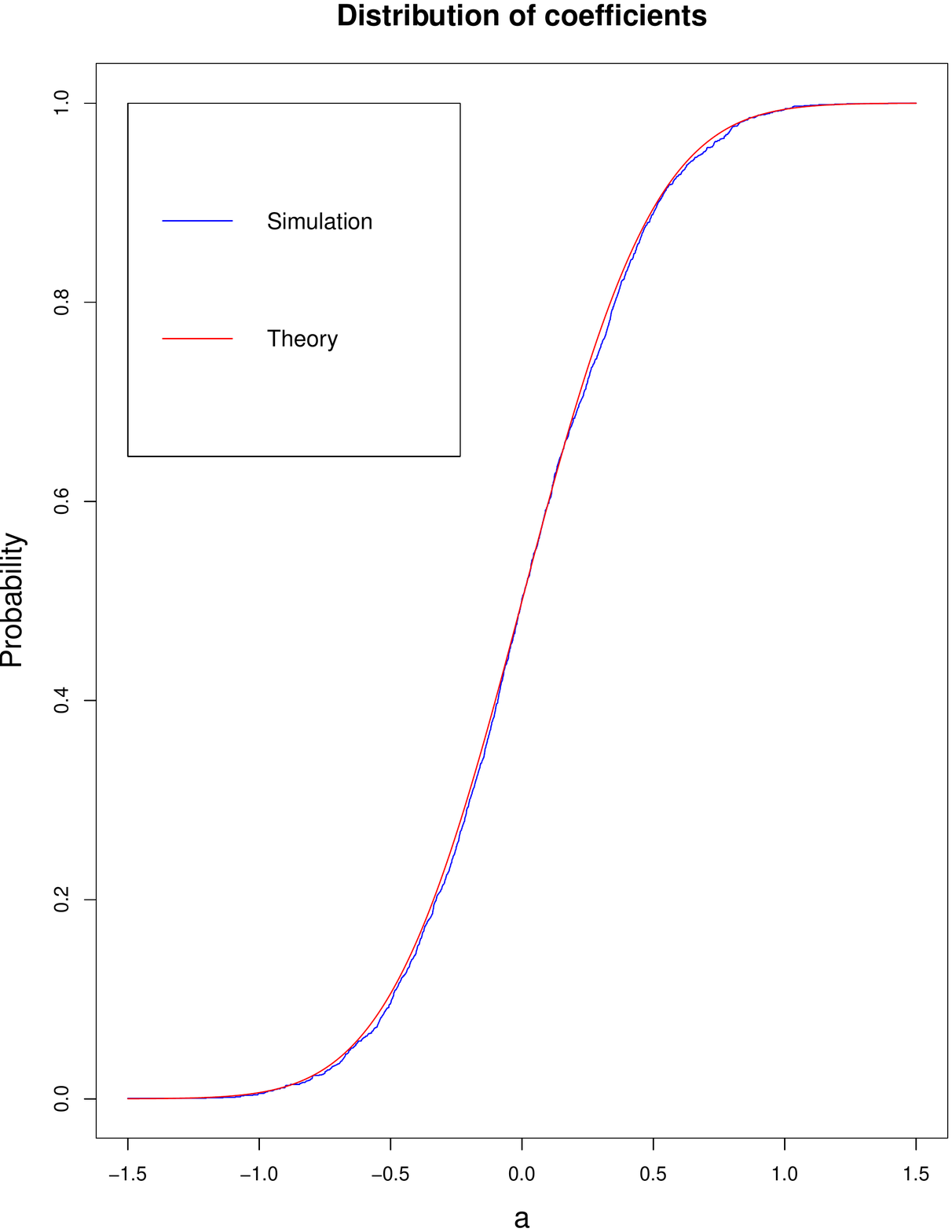}
	\includegraphics[width=5cm,height=5cm]{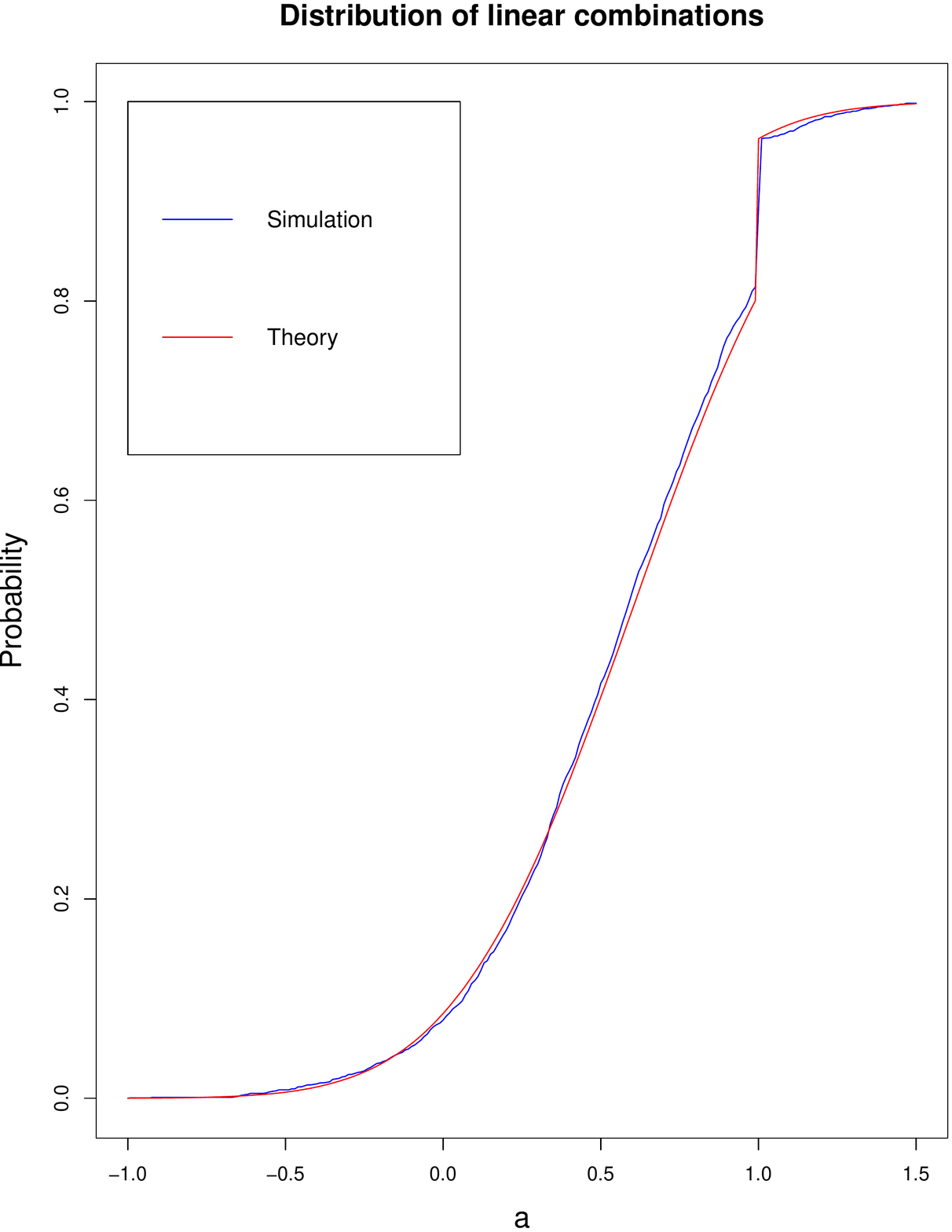}
	\caption{The empirical cumulative distribution against the theoretic cumulative distribution. For the first plot, the blue line is the empirical distribution of $a(j)$'s in simulation and red line is the theoretic prediction. For the second plot, the blue line is the empirical distribution of $L_i$'s obtained in simulation and red line is the theoretic prediction.}
	\label{fig:signal12}
\end{figure}

Here we consider a logistic model:
\[\Prob(y=1|x)=\frac{1}{1+\exp\left(-3x^Ta_0/\sqrt{p}\right)},\]
with a direction $a_0$ generated uniformly at random such that $\norm{a_0}_2^2=p$. We generate data with sample size $n=2000$ and $p=2000$. This corresponds to $\delta=p/n=1$. The SVM is run at $\lambda=1$. We first solve our analytic equations~(\ref{eq:mainsignal}) and (\ref{eq:alpha}). The landscape of the minimization problem in equation~(\ref{eq:alpha}) is shown in the first plot of figure~\ref{fig:together}. And indeed we have approximately $\alpha_*=0.28$. Given the value of $\alpha_*$, solving equation~(\ref{eq:mainsignal}) gives approximately $\gamma_*=0.42,\sigma_*=0.40$. So overall the solution for the analytic equations~(\ref{eq:mainsignal}) and \ref{eq:alpha} is
\[\alpha_*=0.28,\gamma_*=0.42,\sigma_*=0.40.\]
\noindent Then the results in Section~\ref{sec:linearsignal} predicts the following:
\begin{itemize}
	\item The $\ha(j)$'s fluctuate around $\alpha_*a_0(j)$'s.
	\item The centered coefficients $\ha(j)-\alpha_*a_0(j)$'s follow a $N(0,\sigma_*^2)$ distribution.
	\item The linear combinations $L_i=\frac{x_i^T\ha}{\sqrt{p}}$'s approximately follow the same distribution as $f_{\gamma_*}(\alpha_*V+\sigma_* Z)$.
	\item The misclassification error on new data points is approximately $\Prob(\alpha_*V+\sigma_* Z<0)$.
\end{itemize}
The first assertion is verified in the second plot of figure~\ref{fig:together}, where we plot the value of $\ha-\alpha_*a_0$ against the value of $a_0$. The second and third assertion is verified in figure~\ref{fig:signal12}, where the empirical cumulative distribution function and the theoretic cumulative distribution function are plotted against each other. Again, the simulation result is in perfect accordance with all the predictions given by the analytic formulas. Finally, table~\ref{tab:table1} shows the actual misclassification error of $\hat y=\textnormal{sign}(\ha^Tx)$ on new data points and the value predicted by our theory.
\begin{table}[H]
	\begin{center}
		\caption{Misclassification error on new data}
		\label{tab:table1}
		\begin{tabular}{c|c} 
			Empirical value & Theoretical prediction\\
			$0.33$ & $0.34$ \\
		\end{tabular}
	\end{center}
\end{table}

\subsubsection{The indicator model}
\begin{figure}[H]
	\centering
	\includegraphics[width=6cm,height=6cm]{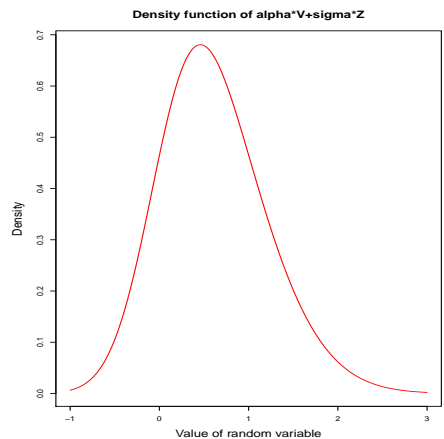}
	\caption{Density function of $\alpha_*V+\sigma_*Z$.}
	\label{fig:indicator}
\end{figure}

Now we consider the indicator model:
\[\Prob(y=1|x)=\indc{a_0^Tx\geq 0},\]
which is just the deterministic relation $y=\textnormal{sign}(a_0^Tx)$.
In our setting, this model corresponds to the extreme case where the signal is the strongest. In this case, the two classes are always exactly separable regardless of the value of $\delta$. Notice that even in this case, the SVM optimization is still well-posed because of the penalty term on the scaling. And in specific, $\ha$ will not align perfectly with the true direction $a_0$ even if the data points are exactly separable. We generate our data with $p=1000,n=4000$ which corresponds to $\delta=0.25$. We run SVM with $\lambda=1$. Solving equation~(\ref{eq:mainsignal}) and (\ref{eq:alpha}) gives approximately:
\[\alpha_*=0.76, \gamma_*=0.24, \sigma_*=0.40.\]
The behavior predicted by the set of solutions is in perfect accordance to the simulation result. To avoid repetition, we show in figure~\ref{fig:indicator} the density function of $\alpha_*V+\sigma_*Z$. We emphasize again all the information one could reads off this density plot: integrating it from $-\infty$ to $0$ gives the misclassification error on new data points; Integrating it from $1-\gamma$ to $1$ gives the proportion of data points on the margin boundary; Integrating it from $-\infty$ to $1-\gamma$ gives the proportion of data violating the margin condition in the training data.

\subsection*{Acknowledgements}
The author is grateful to Rina Foygel Barber for helpful discussions on this work.

\bibliographystyle{plainnat}
\bibliography{main}

\appendix

\section{A derivation of the equations in Section~\ref{sec:globalnull} and Section~\ref{sec:linearsignal}}
Here we give a heuristic derivation of the equations in Section~\ref{sec:globalnull} and Section~\ref{sec:linearsignal}. We start with the global null case. Then the differences in the signaled case are briefly mentioned. Throughout the derivation we use $p(\cdot)$ to denote a general density function, whose specific meaning depends on the context. In the global null case $yx$ still follow a standard normal distribution $N(0,I)$. Therefore we only have to consider the following minimization problem:
\[\ha=\arg\min_a \sum_{i=1}^n \brac{1-\frac{x_i^Ta}{\sqrt{p}}}_++\lambda \sum_{j=1}^p a^2(j).\]
Define $L_i$ to be the linear combination $L_i=\frac{x_i^a}{\sqrt{p}}$. Also let $\eta_K$ be a sequence of smooth function that approximate the function $(1-x)_+$ when $K\rightarrow \infty$. The specific form of $\eta_K$ doesn't matter. But for concreteness, let us set
\[\eta_K=\frac{\log\left(1+\exp(K(x-1))\right)}{k}-(x-1), \eta_K'=\frac{-1}{1+\exp\brac{K(x-1)}}, \eta_K''=\frac{K\exp\brac{K(x-1)}}{\brac{1+\exp\brac{K(x-1)}}^2}.\]
Then we consider the following distribution on $a$:
\[p(a)\propto \exp\brac{-\beta\brac{\sum_{i=1}^n\eta_K(L_i)+\lambda\sum_{j=1}^pa^2(j)}}.\]
Throughout the derivation, we use $\avg{\cdot}$ to denote an average over the randomness of $a$ following the above distribution conditional on the realization of $\{x_1,\dots,x_n\}$, and we use an overline $\overline{\cdot}$ to denote an average over the randomness of $\{x_1,\dots,x_n\}$. We first consider adding a new data point $x_0$ to the existing system, which corresponds to a new linear combination $L_0=\frac{x_0^Ta}{\sqrt{p}}$. With respect to the randomness of $a$ in the old $n$-system, we have
\[L_0\sim N(h,q_0-q_1); h=\frac{\sum_jx_0(j)\avg{a(j)}}{\sqrt{p}},q_0=\overline{\avg{a^2(1)}}, q_1=\overline{\avg{a(1)}^2}.\]
Then with respect to the randomness of $a$ in the new $n+1$-system, the density function of $L_0$ is proportional to
\begin{equation}\label{eq:app1}
p(L)\propto \exp\brac{-\frac{(L-h)^2+2\beta(q_0-q_1)\eta_K(L)}{2(q_0-q_1)}}.
\end{equation}
Denote $\gamma=\beta(q_0-q_1)$. Then when $\beta$ is large this distribution approximately has mean $\prox_{\gamma\eta_K}(h)$ and variance $\frac{q_0-q_1}{1+\gamma\eta_K''\brac{\prox_{\gamma\eta_K}(h)}}$, where we define the proximal operator as  
\[\prox_{\gamma\eta_K}(h)=\arg \min_{\tilde h} \frac{1}{2}(\tilde h-h)^2+\gamma\eta_K(\tilde h).\] 
Moreover with respect to the random of $\{x_0,x_1,\dots,x_n\}$ we have $h\sim N(0,q_1)$. Now we consider adding one dimension $a(0)$ to $a$ and correspondingly add one dimension $x_i(0)$ to each data point $x_i$. Then similar calculation as above shows that with respect to the randomness of $a$ in the new $p+1$ system, we have approximately
\begin{equation}\label{eq:app2}
a(0)\sim N\left(\frac{-\delta\tilde h}{2\lambda\delta+M-\beta(\tilde q_0-\tilde q_1)},\frac{\delta}{\beta(2\lambda\delta+M-\beta(\tilde q_0-\tilde q_1))}\right),
\end{equation}
where
\[\tilde h=\frac{\sum_i x_i(0)\avg{\eta_K'(L_i)}}{\sqrt{p}},\tilde q_0=\overline{\avg{\eta_K'(L_1)^2}}, \tilde q_1=\overline{\avg{\eta_K'(L_1)}^2}, M=\overline{\avg{\eta_K''(L_1)}}.\]
Moreover with respect to the randomness of $\{x_1,\dots,x_n\}$ we have $\tilde h\sim N(0,\tilde q_1/\delta)$. Recall that we denote $\gamma=\beta(q_0-q_1)$. Further define $\sigma$ through $\sigma^2=q_1$. Then equation~(\ref{eq:app1}) and (\ref{eq:app2}) together imply a set of self-consistency equations which in the large $\beta$ limit becomes
\begin{align*}
&\frac{\delta\sigma^2}{\gamma^2}=\E_{h\sim N(0,\sigma^2)}\eta_K'\brac{\prox_{\gamma\eta_K}(h)}^2\\
&(2\lambda\gamma-1)\delta+1=\E_{h\sim N(0,\sigma^2)}\frac{1}{1+\gamma\eta_K''\brac{\prox_{\gamma\eta_K}(h)}}.
\end{align*}
The $K\rightarrow \infty$ limit is then taken which gives rise to
\begin{align*}
&\frac{\sigma^2\delta}{\gamma^2}=\Prob(\sigma Z\leq 1-\gamma)+\E\left(\frac{1-\sigma Z}{\gamma}\right)^2\indc{1-\gamma\leq \sigma Z \leq 1}.\\
&(2\lambda\gamma-1)\delta+1=\Prob(\sigma Z\leq 1-\gamma)+\Prob(\sigma Z\geq 1).
\end{align*}
In the signaled case, since the system is rotational symmetric, we will assume $a_0=e_1$ without loss of generality. The only thing that changes in the above derivation is that when adding a new data point, the distribution of $L_0$ with respect to the randomness of $a$ in the $n$-system will instead have mean value $\alpha y_0x_0(1)+h$. Then with respect to the randomness in $\{x_0,x_1,\dots,x_n\}$, $\alpha V+h$ follows the same distribution of $\alpha V+\sigma^2Z$, where $V$ has the same distribution as $yx(1)$ and $Z\sim N(0,1)$ and they are independent. This give rise to equation~(\ref{eq:mainsignal}) in the signaled case. Equation~(\ref{eq:alpha}), which gives the optimal value of $\alpha_*$, is simply by the definition of the minimization problem of SVM.

\end{document}